\documentclass{article}
\usepackage{nips15submit_e,times}
\usepackage{hyperref}
\usepackage{algorithm}
\usepackage{algorithmic}
\usepackage{amsmath,amsthm,amssymb}
\usepackage{url}
\usepackage{dsfont}
\usepackage{graphicx}

\DeclareMathOperator{\spn}{span}

\newcommand{\comment}[1]{}
\newcommand{\inprod}[2]{\langle #1,#2 \rangle}
\newcommand{\us}{\mathcal{U}}
\newcommand{\R}{\mathbb{R}}
\newcommand{\Bx}{\mathcal{B}_x}

\title{Understanding Adversarial Training: Increasing Local Stability of Neural Nets through Robust Optimization}

\author{
Uri Shaham \\
Yale University\\
\texttt{uri.shaham@yale.edu} \\
\And
Yutaro Yamada \\
Yale University \\
\texttt{yutaro.yamada@yale.edu}
\And
Sahand Negahban \\
Yale University \\
\texttt{sahand.negahban@yale.edu}
}

\nipsfinalcopy 

\begin{document}

\maketitle

\begin{abstract} 
We propose a general framework for increasing local stability of Artificial Neural Nets (ANNs) using Robust Optimization (RO). We achieve this through an alternating minimization-maximization  procedure, in which the loss of the network is minimized over perturbed examples that are generated at each parameter update. We show that adversarial training of ANNs is in fact robustification of the network optimization, and that our proposed framework generalizes previous approaches for increasing local stability of ANNs. Experimental results reveal that our approach increases the robustness of the network to existing adversarial examples, while making it harder to generate new ones. Furthermore, our algorithm improves the accuracy of the network also on the original test data.
\end{abstract} 

\section{Introduction}
\label{introduction}

The fact that ANNs might by very unstable locally was demonstrated in
\cite{szegedy2013intriguing}, where it was shown that highly
performing vision ANNs mis-classify examples that have only barely
perceivable (by a human eye) differences from correctly classified
examples. Such examples are called \textit{adversarial examples}, and
were originally found by solving an optimization problem, with respect to a
trained net. Adversarial examples do not tend to exist naturally in
training and test data. Yet, the local instability manifested by their
existence is somewhat disturbing. In the case of visual data, for
example, one would expect that images that are very close in the
natural ``human eye'' metric will be mapped to nearby points in the
hidden representation spaces, and consequently will be predicted to
the same class. Moreover, it has been shown that different models with different architectures
which are trained on different training sets tend to mis-classify the
same adversarial examples in a similar fashion.  In addition to the
disturbing existence of adversarial examples from model stability
perspective, the fact that they can be generated by simple and
structured procedures and are common to different models can be used
to perform attacks on models by making them fail easily and
consistently \cite{gu2014towards}.

It has been claimed that adversarial
examples exist in 'blind spots' in the data domain in which training and
testing data do not occur naturally; however, some of these blind
spots might be very close in some sense to naturally occurring data.

Several works have proposed to use adversarial examples during training of ANNs, and reported increase in classification accuracy on test data (for example, \cite{szegedy2013intriguing}, \cite{goodfellow2014explaining}). The goal of this manuscript is to provide a framework that yields a full theoretical understanding of adversarial training, as well as new optimization schemes, based on robust optimization. Specifically, we show that generating and using adversarial examples during training of ANNs can be derived from the powerful notion of Robust Optimization (RO), which has many applications in machine learning and is closely related to regularization. We propose a general algorithm for robustification of ANN training, and show that it generalizes previously proposed approaches. 

Essentially, our algorithm increases the stability of ANNs with respect to perturbations in the input data, through an iterative
minimization-maximization procedure, in which the network parameters are
updated with respect to worst-case data, rather than to the original
training data.
Furthermore, we show connections between our method and existing methods for generating adversarial examples and adversarial training, demonstrating that those methods are special instances of the robust optimization framework. This point yields a principled connection highlighting the fact that the existing adversarial training methods aim to robustify the parameter optimization process.

The structure of this paper is as follows: in Section
\ref{sec:related} we mention some of the recent works that analyze
adversarial examples and attempt to improve local stability. In
Section \ref{sec:RO} we present the basic ideas behind Robust Optimization, and some of
its connections to regularization in machine learning models. In
Section \ref{adversarialTraining} we present our training framework,
some of its possible variants and its practical version. Experimental
results are given in Section \ref{sec:experimental}. Section
\ref{sec:conclusions} briefly concludes this manuscript.

\subsection {Notation}
We denote a labeled training set by $\{(x_i,y_i)\}_{i=1}^m$ where
$x_i \in \mathbb{R}^d$ is a set of features and $y_i \in \{1,..,K\}$
is a label. The loss of a network with parameters $\theta$ on $(x,y)$
is denoted by $J(\theta, x,y)$ and is a function that quantifies the
goodness-of-fit between the parameters $\theta$ and the observations
$(x,y)$. When holding $\theta$ and $y$ fixed and viewing
$J(\theta, x,y)$ as a function of $x$ we occasionally write
$J_{\theta,y}( x)$.  $\Delta_x \in \mathbb{R}^d$ corresponds to a
small additive adversarial perturbation, that is to be added to
$x$. By \textit{adversarial example} we refer to the perturbed
example, i.e., $\tilde{x}_i = x + \Delta_x$, along with the original
label $y$. We denote the $\ell_p$ norm for $1 \leq p < \infty$ to be
$\|x\|_p^p = \sum_{j=1}^d |x(i)|^p$ and denote the $\ell_\infty$ norm
of a vector $x$ to be $\|x\|_\infty = \max_i \{|x(i)|\}$. Given two
vectors $x$ and $y$ the Euclidean inner-product is denoted
$\inprod{x}{y} = x^T y = \sum_i x_i y_i$. Given a function $f(x,y)$,
we denote $\nabla_x f(x,y)$ to be the gradient of $f$ with respect to
the vector $x$.

\section{Related Work}
\label{sec:related}
Adversarial examples were first introduced in
\cite{szegedy2013intriguing}, where they were generated for a given
training point $(x,y)$ by using L-BFGS (see ~\cite{wright1999numerical}, for example) to solve the
box-constrained optimization problem
\begin{align}
\label{eq:szegedy}
&\min_{\Delta_x} c\|\Delta_x\|_2 + J(\theta, x + \Delta_x, y') \\
&\mbox{ subject to } x+\Delta_x \in [0,1]^d, \notag
\end{align}
and $y' \neq y$. A similar approach for generating adversarial
examples was used also in \cite{gu2014towards}. The fundamental idea
here is to construct a small perturbation of the data point $x$ in order to force the method to mis-classify the training example $x$ with some incorrect label $y'$.

In \cite{goodfellow2014explaining}, the authors point out that when
the dimension $d$ is large, changing each entry of $x$ by a small
value $\epsilon$ yields a perturbation $\Delta_x$ (such that
$\|\Delta_x\|_\infty = \epsilon$), which can significantly change the
inner-product product $w^Tx$ of $x$ with a weight vector $w$. Their result then chooses to set
the adversarial perturbation
\begin{equation}
  \Delta_x = \epsilon \mbox{sign}(\nabla_x J(\theta,x,y)).\label{eq:eta}
\end{equation}
An alternative formulation of the problem naturally shows how the
adversarial perturbation $\Delta_x$ was obtained. If we take a first-order
approximation of the loss function around the true training example
$x$ with a small perturbation $\Delta_x$
\begin{equation*}
J_{\theta,y}(x+\Delta_x) \approx J_{\theta,y}(x) + \langle\nabla J_{\theta,y}(x), \Delta_x \rangle,
\end{equation*}
and maximize the right hand size with respect to $\Delta_x$ restricted
to an $\ell_\infty$ ball of radius $\epsilon$ we have that the choice
that maximizes the right-hand side is exactly the quantity in
equation~\ref{eq:eta}. Since the gradient $\nabla_x J(\theta,x,y)$ can
be computed efficiently using backpropagation (see, for example, \cite{rojas1996neural}), this
approach for generating adversarial examples is rather fast. In the
sequel we will show how the above computation in an example of the
framework that we present here.

It is reported in \cite {szegedy2013intriguing,
  goodfellow2014explaining} that adversarial examples that were
generated for a specific network were mis-classified in a similar
fashion by other networks, with possibly different architectures and
using different subsets of the data for training.
In \cite{goodfellow2014explaining} the authors claim that this phenomenon
is related to the strong linear nature that neural networks
have. Specifically, they claim that the different models learned by
these neural nets were essentially all close to the same linear model,
hence giving similar predictions on adversarial examples.
 
Two works \cite{nguyen2014deep,fawzi2015analysis}
nicely demonstrate that classifiers can achieve very high test
accuracy without actually learning the true concepts of the classes
they predict. Rather, they can base their predictions on
discriminative information, which suffices to obtain accurate predictions on test data, however does not reflect learning of the true concept that defines specific classes. As a result, they can consistently fail in recognizing the true class concept in new examples \cite{fawzi2015analysis} or confidently give wrong predictions on specifically designed examples \cite{nguyen2014deep}.
 
Several papers propose training procedures and objective functions
designed to make the function computed by the ANN change more slowly
near training and test points. In \cite{szegedy2013intriguing} adversarial examples were generated and fed back to the training set. This procedure is reported to increase the classification accuracy on test data.
In \cite{goodfellow2014explaining} the following loss function is proposed:
\begin{equation}
\tilde{J}(\theta,x,y) = \alpha J(\theta,x,y) + (1-\alpha)J(\theta,x+\Delta_x,y) ,\label{eq:advTrComp}
\end{equation}
with $\Delta_x$ as in equation (\ref{eq:eta}).
The authors report that the resulting net had improved test set accuracy, and also had better performance on new adversarial examples. They further give intuitive explanations of this training procedure being an adversary game, and a min-max optimization over $\ell_\infty$ balls. In this manuscript, we attempt to make the second interpretation rigor, by deriving a similar training procedure from RO framework.

In~\cite{miyato2015distributional} adversarial training is performed without requiring knowledge of the true label. Rather, the loss function contains a term computing the Kullback-Leibler divergence between the predicted distributions of the label, (e.g., softmax) i.e. $KL(p(y|x)||p(y|x+\Delta_x))$.

In \cite{kereliuk2015deep} adversarial examples were generated for music data. The authors report that back-feeding of adversarial examples to the training set did not result in an improved resistance to adversarial examples.

In \cite{rifai2011manifold}, the authors first pre-train each layer of
the network as a contractive autoencoder~\cite{rifai2011contractive},
which, assuming that the data concentrates near a lower dimensional
manifold, penalizes the Jacobian of the encoder at every training point
$x$, so that the encoding changes only in the directions tangent to
the manifold at $x$.  The authors further assume that points belonging
to different classes tend to concentrate near different sub-manifolds,
separated by low density areas.  Consequently, they encourage the
output of a classification network to be constant near every training
point, by penalizing the dot product of the network's gradient with
the basis vectors of the plain that is tangent to the data manifold at
every training point. This is done by using the loss function
\begin{equation}
\tilde{J}(\theta,x,y) = J(\theta,x,y) + \beta\sum_{u \in \mathcal{B}_x} (\langle u,\nabla _x o(x)  \rangle)^2 ,\label{eq:manifoldTangent}
\end{equation}
where $o(x)$ is the output of the network at $x$ and $\mathcal{B}_x$
is the basis of the hyperplane that is tangent to the data manifold at
$x$. In the sequel we will show how this is related to the approach in this manuscript.

The contractive autoencoder loss is also used in \cite{gu2014towards},
where the authors propose to increase the robustness of an ANN via minimization of a loss
function which contains a term that penalizes the Jacobians of the
function computed by each layer with respect to the previous layer.

In \cite{szegedy2013intriguing}, the authors propose to regularize
ANNs by penalizing the operator norm of the weight matrix of every
layer. Such thing which will lead to pushing the Lipschitz constant of
the function computed by a layer down, so that small perturbations in
input will not result in large perturbations in output. We are not
aware of any empirical result using this approach.

A different approach is taken in \cite{oyallon2014deep}, using
scattering convolutional networks (convnets), having wavelets based
filters. The filters are fixed, i.e., not learned from the data, and
are designed to obtain stability under rotations and translations. The
learned representation is claimed to be stable also under small
deformations created by additive noise. However, the performance of
the network is often inferior to standard convnets, which are trained
in a supervised fashion.

Interesting theoretical arguments are presented in
\cite{fawzi2015analysis}, where it is shown that robustness of any
classifier to adversarial examples depends on the
\textit{distinguishability} between the classes; they show that
sufficiently large distinguishability is a necessary condition for
\textit{any} classifier to be robust to adversarial
perturbations. Distinguishability is expressed, for example, by
distance between means in case of linear classifiers and and between
covariance matrices in the case of quadratic classifiers.

\section{Robust Optimization}
\label{sec:RO}

Solutions to optimization problems can be very sensitive to small
perturbations in the input data of the optimization problem, in the
sense that an optimal solution given the current data may turn into a
highly sub-optimal or even infeasible solution given a slight change
in the data.  A desirable property of an optimal solution is to remain
nearly optimal under small perturbations of the data. Since
measurement data is typically precision-limited and might contain
errors, the requirement for a solution to be stable to input
perturbations becomes essential.

\textit {Robust Optimization} (RO, see, for example~\cite{ben2009robust}) is an area of optimization theory
which aims to obtain solutions which are stable under some level of
uncertainty the data. The uncertainty has a deterministic and
worst-case nature. The assumption is that the perturbations to the
data can be drawn from specific sets $\mathcal{U}_i$ called
\textit {uncertainty sets}. The uncertainty sets are often defined in
terms of the type of the uncertainty and a parameter controlling the
size of the uncertainty set. The Cartesian product of the sets $\mathcal{U}_i$ is usually denoted by $\mathcal{U}$.

The goal in Robust Optimization is to obtain solutions which are feasible and
well-behaved under \textit{any} realization of the uncertainty from
$\mathcal{U}$; among feasible solutions, an optimal one would be such
that has the minimal cost given the worst-case realization from
$\mathcal{U}$.  Robust Optimization problems thus usually have a min-max formulation,
in which the objective function is being minimized with respect to a
worst-case realization of a perturbation.  For example, consider
standard linear programming problem
\begin{equation}
\min_x \{c^Tx: Ax \le b \}.
\end{equation}
The given data in this case is $(A,b,c)$ and the goal is to obtain a
solution $x$ which is robust to perturbations in the data. Clearly, no
solution can be well-behaved if the perturbations of the data can be
arbitrary. Hence, we restrict ourselves to only allowing the
perturbations to exist in in the uncertainty set $\mathcal{U}$.  The
corresponding Robust Optimization formulation is
\begin{equation}
\min_x \sup _{(A,b,c) \in \mathcal{U}}\{c^Tx: Ax \le b \}.
\end{equation}
Thus, the goal of the above problem is to pick an $x$ that can work
well for \emph{all} possible instances of the problem parameters
within the uncertainty set. 

The robust counterpart of an optimization problem can sometimes be more complicated to solve than the original problem. \cite{mutapcic2009cutting} and \cite{ben2015oracle} propose algorithms for approximately solving the robust problem, which are based only on the algorithm for the original problem. This approach is closely related to the algorithm we propose in this manuscript.  

In the next section we discuss the
connection between Robust Optimization and
regularization. Regularization serves an important role in Deep
Learning architectures, with methods such as dropout~\cite{wager2013dropout} and
sparsification (for example, \cite{bengio2013representation}) serving as a few examples.
\subsection {Robust Optimization and Regularization}
Robust Optimization is applied in various settings in statistics and
machine learning, including, for example, several parameter estimation
applications.  In particular, there is a strong
connection between Robust Optimization and regularization; in
several cases it was shown that solving a regularized problem is
equivalent to obtaining a Robust Optimization solution for a
non-regularized problem. For example, it was shown in
\cite{xu2009robust} that a solution to a $\ell_1$ regularized least
squares problem
\begin {equation}
\min_x \|Ax-b \| + \lambda \|x \|_1
\end{equation}
is also a solution to the Robust Optimization problem 
\begin{equation}
\min_x \max_{\| \Delta A|_{\infty,2} \le \rho} \|(A + \Delta A )x - b\|,
\end{equation}
where $\| \cdot \|_{\infty,2}$ is the $\ell_\infty$ norm of the
$\ell_2$ norms of the columns \cite{bertsimas2011theory}. As a result,
it was shown~\cite{xu2009robust} that sparsity of the solution
$x^{opt}$ is a consequence of its robustness. Regularized Support
Vector Machines (SVMs) were also shown to have robustness properties:
in \cite{xu2009robustness} it was shown that solutions to SVM with
norm regularization can be obtained from non-regularized Robust
Optimization problems \cite{bertsimas2011theory}. Finally, Ridge
Regression can also be viewed as a variant of a robust optimization
problem. Namely, it can be shown that
\begin{equation*}
  \min_x \max_{\{\Delta : \|\Delta\|_F \leq \gamma\}} \|(A+\Delta) x - b \|_2
\end{equation*}
is equivalent to $\min_x \|A x - b \|_2 + \gamma \|x\|_2$~\cite{sra2012optimization}.

\comment{
In addition, it was shown in \cite{el1997robust} that a solution to
the regularized least squares problem
\begin{equation}
\min_x \|Ax-b\| + \rho \sqrt{\|x\|_2^2+1}
\end{equation}
can also be obtained from solving a corresponding Robust Optimization problem.}
%
%
\section{The Proposed Training Framework}
\label{adversarialTraining}
Inspired by the Robust Optimization paradigm, we propose a loss
function for training ANNs.  Our approach is designed to make
the network's output stable in a small neighborhood around every
training point $x_i$; this neighborhood corresponds to the uncertainty
set $\mathcal{U}_i$. For example, we may set $\us_i = B_\rho(x_i,r)$, a ball with radius $r$
around $x_i$ with respect to some norm $\rho$. To do so, we select from this neighborhood a
representative $\tilde{x}_i = x_i + \Delta_{x_i}$, which is the point
on which the network's output will induce the greatest loss; we then require the network's
output on $\tilde{x}_i$ to be $y_i$, the target output for
$x_i$. Assuming that many test points are indeed close to training
points from the same class, we expect that this training
algorithm will have a regularization effect and consequently will
improve the network's performance on test data. 
Furthermore, since adversarial examples are typically generated in proximity to training
or test points, we expect this approach to increase the robustness of
the network's output to adversarial examples.

We propose training the network using a minimization-maximization approach to optimize:
\begin{equation}
\min_\theta \tilde{J}(\theta,x,y) = \min_\theta \sum_{i=1}^m \max_{\tilde{x}_i \in \us_i} J(\theta, \tilde{x}_i,y_i), \label{eq:advLoss}
\end{equation}
where $\us_i$ is the uncertainty set corresponding to example $i$. This can be viewed as
optimizing the network parameters $\theta$ with respect to a
worst-case data $\{(\tilde{x}_i, y_i )\}$, rather than to the original
training data; the $i$'th worst-case data point is chosen from the uncertainty set
$\us_i$. The uncertainty sets are determined by the type of uncertainty and can be
selected based on the problem at hand. 

Optimization of (\ref{eq:advLoss}) can be done in a standard iterative
fashion, where in each iteration of the algorithm two optimization
sub-procedures are performed. First, the network parameters $\theta$
are held fixed and for every training example $x_i$ an additive
adversarial perturbation $\Delta_{x_i}$ is selected such that $x_i + \Delta_{x_i} \in \us_i$ and
\begin{equation}
\Delta_{x_i} = \arg\max_{\Delta : x_i + \Delta \in \us_i} J_{\theta,y_i} (x_i+\Delta) \label{eq:delta}.
\end{equation}
Then, the network parameters $\theta$ are updated with respect to the
perturbed data $\{(\tilde{x}_i,y_i)\}$, where
$\tilde{x}_i = x_i +\Delta_{x_i}$. This maximization is related to
the adversarial example generation process previously proposed by
Szegedy et. al.~\cite{szegedy2013intriguing} as shown in
equation~\eqref{eq:szegedy}.

Clearly, finding the exact $\Delta_{x_i}$ in Equation~\eqref{eq:delta} is
intractable in general. Furthermore, performing a full optimization
process in each of these sub-procedures in each iteration is not
practical. Hence, we propose to minimize a
surrogate to $\tilde{J}$, in which each sub-procedure is reduced to a
single ascent / descent step; that is, in each iteration, we perform a
single ascent step (for each $i$) to find an approximation
$\hat{\Delta}_{x_i}$ for $\Delta_{x_i}$, followed by a single descent
step to update $\theta$. The surrogate that we consider is the
first-order Taylor expansion of the loss around the example, which yields:
\begin{equation}
  \label{eq:upDelx}
  \hat{\Delta}_{x_i} \in \arg \max_{\Delta: x_i + \Delta \in \us_i} J_{\theta,y_i}(x_i) + \langle\nabla J_{\theta,y_i}(x), \Delta \rangle. 
\end{equation}
Our proposed training procedure is formalized in Algorithm
\ref{algo:advTraining}. In words, the algorithm performs
alternating ascent and descent steps, where we first ascend for each $i$ with respect to
the training example $x_i$ and descend with respect to network
parameters $\theta$.

\begin{algorithm}[tb]
   \caption{Adversarial Training}
   \label{alg:example}
\begin{algorithmic}
   \STATE {\bfseries Input:} $\{(x_i,y_i) \}_{i=1}^m   $
   \STATE {\bfseries Output:} robust parameter vector $\theta$
   \STATE initialize $\theta$
   \WHILE{$\theta$ not converged}
      \FOR{every mini batch $mb$}
        \FOR{$i=1,..,|mb|$}
          \STATE Compute $\hat{\Delta}_{x_i} $ using a single ascent step to approximate $\Delta_{x_i}$ via  equation (\ref{eq:upDelx})
          \STATE Set $\tilde{x}_i \leftarrow x_i + \hat{\Delta}_{x_i}$
        \ENDFOR
     
        \STATE Update $\theta$ using a single descent step with respect to the perturbed data $\{(\tilde{x}_i,y_i) \}_{i=1}^{|mb|}$
     \ENDFOR
   \ENDWHILE
\end{algorithmic}
\label{algo:advTraining}
\end{algorithm}

Note that under this procedure, $\theta$ is never updated with respect
to the original training data; rather, it is always updated with
respect to worst-case examples which are close to the original training
points with respect to the uncertainty sets $\us_i$. In the sequel, we will remark on how to solve equation~\eqref{eq:upDelx} for special cases of $\us_i$. In general, one could use an algorithm like L-BFGS or projected gradient descent~\cite{Nesterov04}.

Finally, note that in each iteration of the algorithm two forward and
backward passes through the network are performed, one using the
original training data to compute the adversarial perturbations
$\tilde{x}_i$ and one using the perturbed data to compute the update
for $\theta$; hence, we expect the training time to be twice as long,
comparing to standard training.

\subsection{Examples of uncertainty sets}
There is a number of cases that one can consider for the uncertainty
sets $\us_i$. One example is when $\us_i = B_\rho(x_i,r)$, a norm ball centered at $x_i$ with radius $r$ with respect to the norm $\rho$. Some
interesting choices for $\rho$ are the
$\ell_\infty$, $\ell_1$ and $\ell_2$ norms. Thus, $\Delta_{x_i}$ can then be
approximated using normalized steepest ascent step with respect to the norm
$\rho$~\cite{boyd2004convex}. The steepest ascent step with respect to the $\ell_\infty$ ball
(i.e., box) is obtained by the sign of the gradient
$\mbox{sign} \nabla J_{\theta,y_i}(x_i)$. Choosing $\Delta_{x_i}$ from an
$\ell_\infty$ ball will therefore yield a perturbation in which every entry of $x$
is changed by the same amount $r$. The steepest ascent with respect to
the $\ell_2$ ball coincides with the direction of the gradient
$\nabla J_{\theta,y_i}(x_i)$. \comment{The steepest ascent with respect to
  the $\ell_1$ ball corresponds to the direction $g=(g_1,...,g_d)$, where
  $g_i=\mathds{1}_{[\nabla J_{\theta,y}(x)]_i= \|\nabla
    J_{\theta,y}(x)\|_\infty}$.}
Choosing $\Delta_{x_i}$ from an $\ell_1$ ball will yield a sparse
perturbation, in which only one or a small number of the entries of $x_i$ are
changed (those of largest magnitude in  $\nabla J_{\theta,y_i}(x_i)$). 
Observe that in all three cases the steepest ascent direction is derived from the gradient $\nabla J_{\theta,y_i}(x_i)$, which can be
computed efficiently using backpropagation. 
In Section \ref{sec:experimental} we use each of the $\ell_1$,$\ell_2$,$\ell_\infty$ norms to generate adversarial examples by Equation~\eqref{eq:upDelx} and compare the performance of Algorithm~\ref{algo:advTraining} using each of these types of uncertainty sets.

\subsection {Relation to previous works}
The loss function in equation (\ref{eq:advTrComp}), which is proposed
in \cite {goodfellow2014explaining}, can be viewed as a variant of our
approach, in which $\Delta_{x_i}$ is chosen from an
$\ell_\infty$ ball around $x_i$, since $\theta$ is updated with respect
to adversarial examples generated by equation (\ref{eq:eta}), which is
the steepest ascent step with respect to the $\ell_\infty$ norm. Namely,
we simply see that the solution to equation~\eqref{eq:upDelx} for the
case that $\us_i = B_{\ell_\infty}(x_i,\epsilon)$ is the update presented in
equation~\eqref{eq:eta}.

We may also relate our proposed methodology to the Manifold Tangent
Classifier loss function~\cite{rifai2011manifold}. Following their
assumption, suppose that the data exists on a low-dimension smooth
manifold $\Gamma \subset \R^d$. Let the uncertainty set for training
sample $x$ be $\us = \Gamma \bigcap B_{\ell_2}(x,r)$. Thus, we
would like to obtain the perturbation $\Delta_{x}$ by solving $\Delta_x = \sup_{x+\delta\in \us}J_{\theta,y}(x+\delta)$
\comment{\begin{align*}
\end{align*}}
Again, we take a first-order Taylor approximation of
$J_{\theta,y}(x)$ around $x$ and obtain
$J_{\theta,y}(x+\delta) \approx J_{\theta,y}(x) + \langle \delta,
\nabla_x J_{\theta,y}(x) \rangle$.
We then obtain $\hat{\Delta}_x$ through the optimization
\begin{equation}
  \hat{\Delta}_x = \arg \max_{\Delta:x + \Delta \in \us} J_{\theta,y}(x) + \langle \nabla_x J_{\theta,y}(x), \Delta \rangle \label {eq:smoothD}
\end{equation}
Recalling that $\Gamma$ is locally Euclidean, denoting by
$\mathcal{B}_x$ the basis for the hyperplane that is tangent to
$\Gamma$ at $x$ and given that $r$ is sufficiently small, we may rewrite equation~\eqref{eq:smoothD} as
\begin{equation}
\arg \max_{{\Delta\in \spn\mathcal{B}_x,\, \|\Delta \|_2 \le r}}J_{\theta,y}(x) + \langle \nabla_x J_{\theta,y}(x), \Delta \rangle, 
\end{equation}
The solution to the above equation is
$\Delta_x \propto \Pi_{\Bx} \nabla_x J_{\theta,y}(x)$, where
$\Pi_{\Bx}$ is the orthogonal projection matrix onto the subspace
$\Bx$ and $\Delta_x$ should have $\ell_2$ norm equal to $r$. Thus,
this acts as an $\ell_2$ regularization of the gradient of the loss
with respect to the training sample $x$, projected along the tangent
space $\Bx$, which is analogous to the regularization presented in
equation~\eqref{eq:manifoldTangent}. Put another way, small
perturbations of $x$ on the tangent manifold $\Bx$ should cause very
small changes to the loss, which in turn should result in small perturbations of the output of network
on input $x$.

\section {Experimental Results}
\label{sec:experimental}
In this section we experiment with our proposed training algorithm on
two popular benchmark datasets: MNIST~\cite{lecun1998mnist} and
CIFAR-10~\cite{krizhevsky2010convolutional}. In each case we compare the robustness of a
network that was trained using Algorithm~\ref{algo:advTraining} to that of a
network trained in a standard fashion.

\subsection {Experiments on MNIST dataset} \label{sec:mnistExperiment}
As a baseline, we trained a convnet with ReLU units, two convolutional layers (containing $32$ and $64$ $5 \times 5$ filters), max pooling ($3 \times 3$ and $2 \times 2$ ) after every convolutional layer, and two fully connected layers (of sizes 200 and 10) on top. This convnet had $99.09\%$ accuracy on the MNIST test set. We refer to this network as ``the baseline net''.
 We then used the baseline net to generate a collection of adversarial examples, using equation~\eqref{eq:upDelx}, with $\ell_1,\ell_2$ and $\ell_\infty$ norm balls.  
 
\begin{figure}[ht!]
   \centering 
     \includegraphics[width=2.5in,height=0.8in]{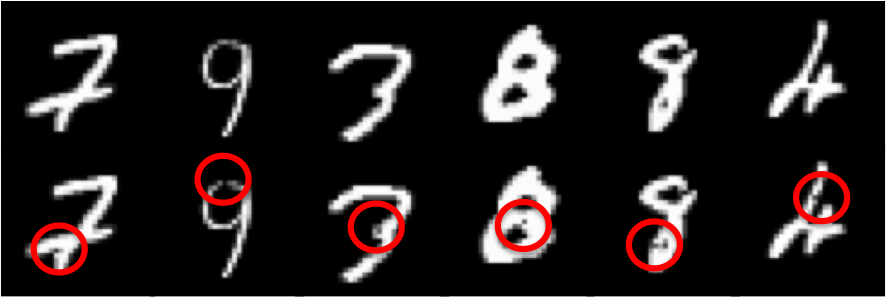}
    \caption{Adversarial examples that were generated for the MNIST dataset w.r.t to the baseline net, via equation~\eqref{eq:upDelx} with $\ell_1$ uncertainty. Top row: original test examples. Bottom row: adversarial examples, where a single pixel (circled) was changed. All original examples presented here were correctly classified by the baseline net, all adversarial examples were mis-classified.}
   \label{fig:adversarialsL1}
 \end{figure}

Specifically, the adversarial perturbation was computed by a step in the steepest ascent direction w.r.t the corresponding norm. The step w.r.t to $\ell_\infty$ uncertainty set is the same as the fast method of \cite{goodfellow2014explaining}; the step w.r.t to $\ell_2$ uncertainty set is in the direction of the gradient; 
the steepest ascent direction w.r.t to $\ell_1$ uncertainty sets comes down to changing the pixel corresponding to the entry of largest magnitude in the gradient vector. It is interesting to note that using equation~\eqref{eq:upDelx} with $\ell_1$ uncertainty, it is possible to make a network mis-classify an image by changing only a \textit{single} pixel. Several such examples are presented in Figure~\ref{fig:adversarialsL1}.

\begin{figure}[ht!]
   \centering 
     \includegraphics[width=2.5in,height=0.8in]{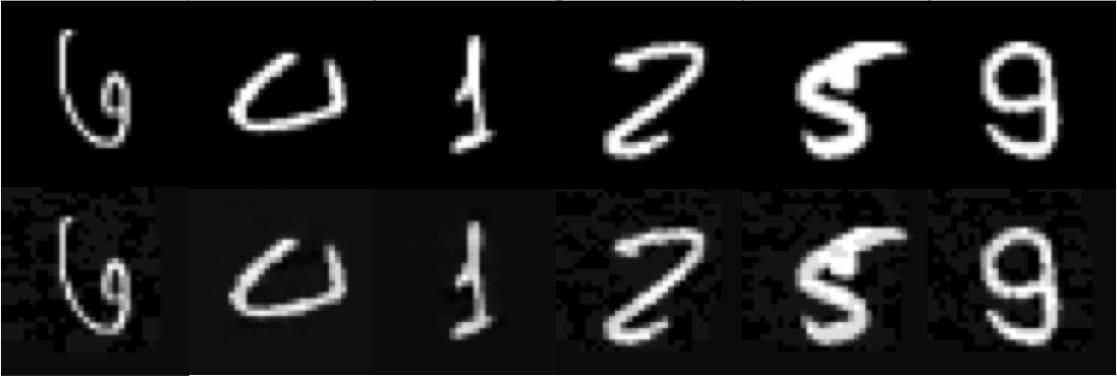}
    \caption{A sample from the set $\mathcal{A}_{\text{mnist}}$ of adversarial examples, generated via equation~\eqref{eq:upDelx}. Top row: original test examples (correctly classified by the baseline net). Bottom row: adversarial examples (mis-classified).}
   \label{fig:adversarialsOrg}
 \end{figure}

 Altogether we generated a collection of 1203 adversarial examples, on which the baseline network had zero accuracy and which were generated from correctly classified test points. A sample of the adversarial examples is presented in Figure \ref{fig:adversarialsOrg}. We refer to this collection as $\mathcal{A}_{\text{mnist}}$.

We then used Algorithm~\ref{algo:advTraining} to re-train the net with the norm $\rho$ being $\ell_1, \ell_2$ and $\ell_\infty$ (each norm in a different experiment). We refer to the resulting nets as the \textit{robustified nets}.
Table~\ref{tab:mnist} summarizes the accuracy of each robustified net on the mnist test data and the collection $\mathcal{A}_{\text{mnist}}$ of adversarial examples.

 \begin{table}[h]
\centering
\caption{Accuracy of the baseline net and each of the three robustified nets on the original MNIST test data, and on the set $\mathcal{A}_{\text{mnist}}$ of adversarial examples that were generated w.r.t to the baseline net.}
\vskip 0.15in
\begin{tabular}{ l || c | r }
  \hline			
  Net & MNIST test set & $\mathcal{A}_{\text{mnist}}$  \\ \hline\hline	
  Baseline & 99.09\% & 0\% \\ \hline\hline	
  Robust $\ell_1$ & 99.16\% & 33.83\% \\ \hline	
  Robust $\ell_2$ & 99.28\% & 76.55\% \\ \hline	
  Robust $\ell_\infty$ & 99.33\% & 79.96\% \\ \hline	
  \end{tabular} 
  
  \label{tab:mnist}
\end{table}

As can be seen, all three robustified nets classify correctly many of the adversarial examples in $\mathcal{A}_{\text{mnist}}$, with the $\ell_\infty$ uncertainty giving the best performance. In addition, all three robustified nets improve the accuracy also on the original test data, i.e., the adversarial training acts as a regularizer, which improves the network's generalization ability. This observation is consistent with the ones in \cite{szegedy2013intriguing} and \cite{goodfellow2014explaining}. 

\begin{figure}[ht!]
   \centering
     \includegraphics[width=3.2in,height=2.5in]{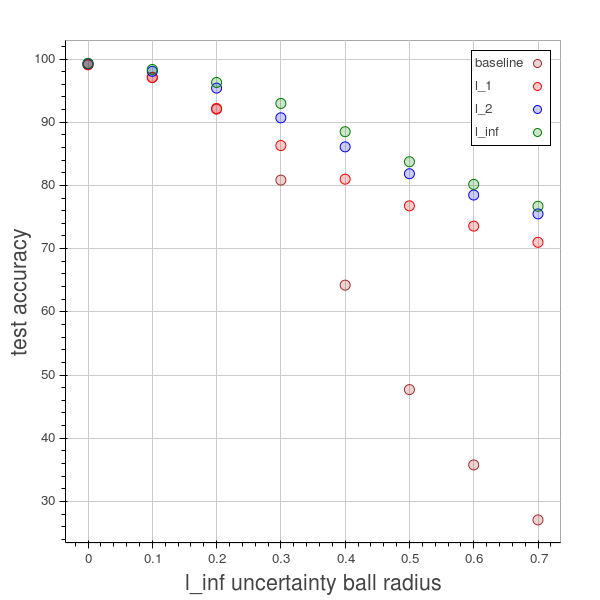}
    \caption{MNIST dataset experiment: comparison between the baseline net (trained in a standard way) and the robustified nets (trained using Algorithm~\ref{algo:advTraining}, with $\ell_1,\,\ell_2$ and $\ell_\infty$ uncertainty sets). Adversarial examples were generated via Equation~\eqref{eq:eta} \textit{with respect to each net} for various values of $\epsilon$ and classification accuracy is plotted. As can be seen, the nets that were trained using Algorithm~\ref{algo:advTraining}) are significantly more robust to adversarial examples.}
   \label{fig:RO}
 \end{figure}

Next, we checked whether it is harder to generate new adversarial examples from the robustified nets (i.e., the nets that were trained via Algorithm~\ref{algo:advTraining}) than from the baseline net. To do that, we used the fast method of \cite{goodfellow2014explaining} (see equation (\ref{eq:eta})) with various values of $\epsilon$ (which corresponds to the amount of noise added/subtracted to/from each pixel) to generate adversarial examples for the baseline net, and for the robustified nets. For each $\epsilon$ we measured the classification accuracy of each net with respect to adversarial examples that were generated from its own parameters. The results 
are shown in Figure \ref{fig:RO}. Clearly, all three robustified nets are significantly more robust to generation of new adversarial examples. 

 To summarize the MNIST experiment, we observed that networks trained with Algorithm~\ref{algo:advTraining} (1) have improved performance on original test data, (2) have improved performance of original adversarial examples that were generated w.r.t to the baseline net, and (3) are more robust to generation of new adversarial examples.
 
%
 \subsection {Experiments on CIFAR-10 dataset}
As a baseline net, we use a variant of the VGG net, publicly available online at~\cite{cifarVGG}, where we disabled the batch-flip module, which flips half of the images in every batch. This baseline net achieved accuracy of 90.79\% on the test set. 

\begin{figure}[ht!]
   \centering
     \includegraphics[width=3in,height=0.8in]{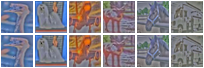}
    \caption{A sample from the set $\mathcal{A}_{\text{cifar10}}$ of adversarial examples.  Some (pre-processed) original CIFAR-10 test examples (top row) and their corresponding adversarial examples (bottom row) for the baseline net. All original test examples are classified correctly by the baseline net while the adversarial examples are all mis-classified. The adversarial examples shown here were generated from the baseline net using equation~\eqref{eq:upDelx} with $\ell_2$ and $\ell_\infty$ uncertainties.}
   \label{fig:cifarAdversarialsOrg}
 \end{figure}

As in section \ref{sec:mnistExperiment}, we constructed adversarial examples for the baseline net, using Equation\eqref{eq:upDelx}, with $\ell_1$, $\ell_2$ and $\ell_\infty$ uncertainty sets. Altogether we constructed 1712 adversarial examples, all of which were mis-classified by the baseline net, and were constructed from correctly classified test images. We denote this set as $\mathcal{A}_{\text{cifar10}}$.
A sample from $\mathcal{A}_{\text{cifar10}}$ is presented in Figure \ref{fig:cifarAdversarialsOrg}.

\begin{figure}[ht!]
   \centering
     \includegraphics[width=3.2in,height=2.5in]{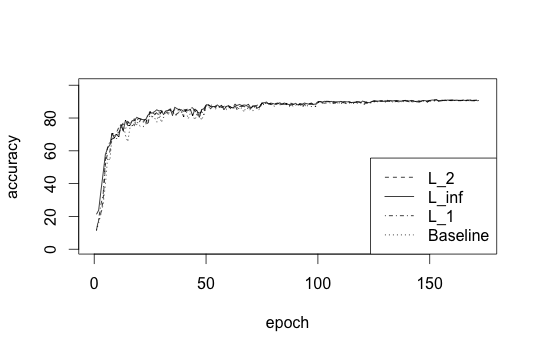}
    \caption{CIFAR-10 dataset experiment: test accuracy vs. epoch number.}
   \label{fig:cifarConvergence}
 \end{figure}

We then used Algorithm~\ref{algo:advTraining} to re-train the net with $\ell_1$, $\ell_2$ and $\ell_\infty$ uncertainty. 
Figure~\ref{fig:cifarConvergence} shows that the robustified nets take about the same number of epochs to converge as the baseline net. 
 
 Table~\ref{tab:cifar} compares the performance of the baseline and robustified nets on the CIFAR-10 test data and the collection $\mathcal{A}_{\text{cifar10}}$ of adversarial examples.
\begin{table}[h]
\centering
\caption{Accuracy of the baseline and the $\ell_2$ and $\ell_\infty$ robustified nets on the original CIFAR-10 test data, and on the set $\mathcal{A}_{\text{cifar10}}$ of adversarial examples that were generated w.r.t to the baseline net.}
\vskip 0.15in
\begin{tabular}{ l || c | r }
  \hline			
  Net & CIFAR-10 test set & $\mathcal{A}_{\text{cifar10}}$  \\ \hline\hline	
  Baseline & 90.79\% & 0\% \\ \hline\hline	
  Robust $\ell_1$ & 91.11\% & 56.31\% \\ \hline	
  Robust $\ell_2$ & 91.04\% & 59.92\% \\ \hline	
  Robust $\ell_\infty$ & 91.36\% & 65.01\% \\ \hline		
  \end{tabular} 
  
  \label{tab:cifar}
\end{table}

Consistently with the results of the MNIST experiment, here as well the robustified nets classify correctly many of the adversarial examples in $\mathcal{A}_{\text{mnist}}$, and also outperform the baseline net on the original test data.

\begin{figure}[ht!]
   \centering
     \includegraphics[width=3.2in,height=2.5in]{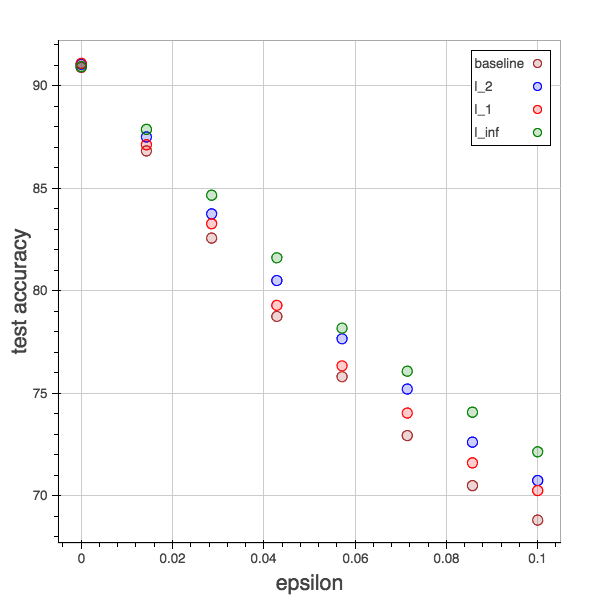}
    \caption{CIFAR-10 dataset experiment: comparison between the baseline net (trained in a standard way) and the robustified nets (trained using Algorithm~\ref{algo:advTraining}, using $\ell_1$, $\ell_2$ and $\ell_\infty$ uncertainty). Adversarial examples were generated via Equation~\eqref{eq:eta} \textit{with respect to each net} for various values of $\epsilon$ and classification accuracy is plotted. The robustified nets are more robust to generation of new adversarial examples.}
   \label{fig:CifarRO}
 \end{figure}

As in the MNIST experiment, we continued by checking whether it is harder to generate new adversarial examples from the robustified nets than from the baseline net; we used equation~\eqref{eq:eta} (i.e., with $\ell_\infty$ uncertainty) and various values of $\epsilon$ to generate adversarial examples for each of the the baseline and the robustified nets. The results are shown in Figure \ref{fig:CifarRO}. 
We can see that new adversarial examples are consistently harder to generate for the robustified net, which is consistent with the observation we had in the MNIST experiment.

 To summarize the CIFAR-10 experiment, we observed that here as well, the robustified nets improve the performance on original test data, while making the nets more robust to generation of new adversarial examples. As in the MNIST experiment, the $\ell_\infty$ uncertainty yields the best improvement in test accuracy. In addition, the robustified nets require about the same number of parameter updates to converge as the baseline net.

\section{Conclusions}
In an attempt to theoretically understand successful empirical results with adversarial training, we proposed a framework for robust optimization of neural nets, in which the network's prediction is encouraged to be consistent in a small ball with respect to some norm. 
The implementation is done using minimization-maximization approach, where the loss is minimized over worst-case examples, rather than on the original data. 
Our framework explains previously reported empirical results, showing that incorporating adversarial examples during training improves accuracy on test data. 
In addition, we showed that the loss function published in \cite{goodfellow2014explaining} is in fact a special case of Algorithm~\ref{algo:advTraining}, for certain type of uncertainty, thus explaining intuitive interpretations given in that paper. We also showed a connection between Algorithm~\ref{algo:advTraining} and the manifold tangent classifier \cite{rifai2011manifold}, showing that it too, corresponds to a robustification of ANN training.

Experimental results in MNIST and CIFAR-10 datasets show that Algorithm~\ref{algo:advTraining} indeed acts as a regularizer and improves the prediction accuracy also on the original test examples, and are consistent with previous results in \cite{goodfellow2014explaining} and \cite{szegedy2013intriguing}.
Furthermore, we showed that new adversarial examples are harder to generate for a network that is trained using our proposed approach, comparing to a network that was trained in a standard fashion. As a by-product, we also showed that one may be able to make a neural net mis-classify a correctly-classified an image by changing only a single pixel.

Explaining the regularization effect that adversarial training has is in the same vein that from practical experience, most authors knew that drop-out (or more generally adding noise to data) acts as regularization, without a formal rigor justification. Later (well-cited) work by Wager, Wang, and Liang \cite{wager2013dropout} created a rigorous connection between dropout and weighted ridge regression. 

The scripts that were used for the experiments are available online at \url{https://github.com/yutaroyamada/RobustTraining} .


\bibliography{adversarial}{}

\begin{thebibliography}{10}

\bibitem{ben2009robust}
Aharon Ben-Tal, Laurent El~Ghaoui, and Arkadi Nemirovski.
\newblock {\em Robust optimization}.
\newblock Princeton University Press, 2009.

\bibitem{ben2015oracle}
Aharon Ben-Tal, Elad Hazan, Tomer Koren, and Shie Mannor.
\newblock Oracle-based robust optimization via online learning.
\newblock {\em Operations Research}, 2015.

\bibitem{bengio2013representation}
Yoshua Bengio, Aaron Courville, and Pascal Vincent.
\newblock Representation learning: A review and new perspectives.
\newblock {\em Pattern Analysis and Machine Intelligence, IEEE Transactions
  on}, 35(8):1798--1828, 2013.

\bibitem{bertsimas2011theory}
Dimitris Bertsimas, David~B Brown, and Constantine Caramanis.
\newblock Theory and applications of robust optimization.
\newblock {\em SIAM review}, 53(3):464--501, 2011.

\bibitem{boyd2004convex}
Stephen Boyd and Lieven Vandenberghe.
\newblock {\em Convex optimization}.
\newblock Cambridge university press, 2004.

\bibitem{fawzi2015analysis}
Alhussein Fawzi, Omar Fawzi, and Pascal Frossard.
\newblock Analysis of classifiers' robustness to adversarial perturbations.
\newblock {\em arXiv preprint arXiv:1502.02590}, 2015.

\bibitem{goodfellow2014explaining}
Ian~J Goodfellow, Jonathon Shlens, and Christian Szegedy.
\newblock Explaining and harnessing adversarial examples.
\newblock {\em arXiv preprint arXiv:1412.6572}, 2014.

\bibitem{gu2014towards}
Shixiang Gu and Luca Rigazio.
\newblock Towards deep neural network architectures robust to adversarial
  examples.
\newblock {\em arXiv preprint arXiv:1412.5068}, 2014.

\bibitem{kereliuk2015deep}
Corey Kereliuk, Bob~L Sturm, and Jan Larsen.
\newblock Deep learning and music adversaries.
\newblock {\em arXiv preprint arXiv:1507.04761}, 2015.

\bibitem{krizhevsky2010convolutional}
Alex Krizhevsky and G~Hinton.
\newblock Convolutional deep belief networks on cifar-10.
\newblock {\em Unpublished manuscript}, 2010.

\bibitem{lecun1998mnist}
Yann LeCun and Corinna Cortes.
\newblock The mnist database of handwritten digits, 1998.

\bibitem{miyato2015distributional}
Takeru Miyato, Shin-ichi Maeda, Masanori Koyama, Ken Nakae, and Shin Ishii.
\newblock Distributional smoothing with virtual adversarial training.
\newblock {\em stat}, 1050:13, 2015.

\bibitem{mutapcic2009cutting}
Almir Mutapcic and Stephen Boyd.
\newblock Cutting-set methods for robust convex optimization with pessimizing
  oracles.
\newblock {\em Optimization Methods \& Software}, 24(3):381--406, 2009.

\bibitem{Nesterov04}
Y.~Nesterov.
\newblock {\em Introductory Lectures on Convex Optimization}.
\newblock Kluwer Academic Publishers, New York, 2004.

\bibitem{nguyen2014deep}
Anh Nguyen, Jason Yosinski, and Jeff Clune.
\newblock Deep neural networks are easily fooled: High confidence predictions
  for unrecognizable images.
\newblock {\em arXiv preprint arXiv:1412.1897}, 2014.

\bibitem{oyallon2014deep}
Edouard Oyallon and St{\'e}phane Mallat.
\newblock Deep roto-translation scattering for object classification.
\newblock {\em arXiv preprint arXiv:1412.8659}, 2014.

\bibitem{rifai2011manifold}
Salah Rifai, Yann~N Dauphin, Pascal Vincent, Yoshua Bengio, and Xavier Muller.
\newblock The manifold tangent classifier.
\newblock In {\em Advances in Neural Information Processing Systems}, pages
  2294--2302, 2011.

\bibitem{rifai2011contractive}
Salah Rifai, Pascal Vincent, Xavier Muller, Xavier Glorot, and Yoshua Bengio.
\newblock Contractive auto-encoders: Explicit invariance during feature
  extraction.
\newblock In {\em Proceedings of the 28th International Conference on Machine
  Learning (ICML-11)}, pages 833--840, 2011.

\bibitem{rojas1996neural}
Ra{\'u}l Rojas.
\newblock {\em Neural networks: a systematic introduction}.
\newblock Springer Science \& Business Media, 1996.

\bibitem{sra2012optimization}
Suvrit Sra, Sebastian Nowozin, and Stephen~J Wright.
\newblock {\em Optimization for machine learning}.
\newblock Mit Press, 2012.

\bibitem{szegedy2013intriguing}
Christian Szegedy, Wojciech Zaremba, Ilya Sutskever, Joan Bruna, Dumitru Erhan,
  Ian Goodfellow, and Rob Fergus.
\newblock Intriguing properties of neural networks.
\newblock {\em arXiv preprint arXiv:1312.6199}, 2013.

\bibitem{wager2013dropout}
Stefan Wager, Sida Wang, and Percy~S Liang.
\newblock Dropout training as adaptive regularization.
\newblock In {\em Advances in Neural Information Processing Systems}, pages
  351--359, 2013.

\bibitem{wright1999numerical}
Stephen~J Wright and Jorge Nocedal.
\newblock {\em Numerical optimization}, volume~2.
\newblock Springer New York, 1999.

\bibitem{xu2009robust}
Huan Xu, Constantine Caramanis, and Shie Mannor.
\newblock Robust regression and lasso.
\newblock In {\em Advances in Neural Information Processing Systems}, pages
  1801--1808, 2009.

\bibitem{xu2009robustness}
Huan Xu, Constantine Caramanis, and Shie Mannor.
\newblock Robustness and regularization of support vector machines.
\newblock {\em The Journal of Machine Learning Research}, 10:1485--1510, 2009.

\bibitem{cifarVGG}
Sergey Zagoruyko.
\newblock 92.45\% on cifar-10 in torch.
\newblock \url{http://torch.ch/blog/2015/07/30/cifar.html}, 2015.
\newblock Online; accessed 12-November-2015.

\end{thebibliography}
\bibliographystyle{plain}

\end{document}